\documentclass[review]{elsarticle}

\usepackage{lineno,hyperref}

\usepackage[letterpaper, margin=1in]{geometry}

\journal{Computers and Electronics in Agriculture}

\biboptions{authoryear}

\usepackage{amssymb}
\usepackage{amsmath}
\usepackage{algorithm}
\usepackage{algpseudocode}
\usepackage{booktabs}
\usepackage{colortbl}
\usepackage{float}
\usepackage{hyperref}
\usepackage{multirow}
\usepackage{natbib}
\usepackage{nicematrix}
\usepackage{siunitx}
\usepackage{subfigure}
\usepackage{tabularx}
\usepackage{xpatch}

\definecolor{Gray}{gray}{0.90}

\DeclareMathOperator*{\argmin}{arg\,min}

\begin{document}

\begin{frontmatter}

\title{Vision-based Navigation of Unmanned Aerial Vehicles in Orchards: An Imitation Learning Approach}


\author[BAE]{Peng Wei}

\author[MAE]{Prabhash Ragbir}

\author[BAE]{Stavros G. Vougioukas}

\author[MAE]{Zhaodan Kong\corref{mycorrespondingauthor}}
\cortext[mycorrespondingauthor]{Corresponding author}
\ead{zdkong@ucdavis.edu}

\address[BAE]{Department of Biological and Agricultural Engineering, University of California, Davis, Davis, CA, USA}
\address[MAE]{Department of Mechanical and Aerospace Engineering, University of California, Davis, Davis, CA, USA}

\begin{abstract}
Autonomous unmanned aerial vehicle (UAV) navigation in orchards presents significant challenges due to obstacles and GPS-deprived environments. In this work, we introduce a learning-based approach to achieve vision-based navigation of UAVs within orchard rows. Our method employs a variational autoencoder (VAE)-based controller, trained with an intervention-based learning framework that allows the UAV to learn a visuomotor policy from human experience. We validate our approach in real orchard environments with a custom-built quadrotor platform. Field experiments demonstrate that after only a few iterations of training, the proposed VAE-based controller can autonomously navigate the UAV based on a front-mounted camera stream. The controller exhibits strong obstacle avoidance performance, achieves longer flying distances with less human assistance, and outperforms existing algorithms. Furthermore, we show that the policy generalizes effectively to novel environments and maintains competitive performance across varying conditions and speeds. This research not only advances UAV autonomy but also holds significant potential for precision agriculture, improving efficiency in orchard monitoring and management.
\end{abstract}


\end{frontmatter}


\section{Introduction}
\label{sec:introduction}

Unmanned aerial vehicle (UAV) technology has made significant progress in recent years, particularly for applications in agriculture. The ability to navigate within orchard rows allows UAVs to perform tasks such as crop inspection and yield estimation~\citep{zhang2021orchard}. This capability provides a valuable tool for remote sensing and precision agriculture~\citep{chen2022predicting}, leading to more efficient and improved orchard management. However, most existing UAVs still depend on GPS for navigation in agricultural settings. This reliance limits their ability to operate in confined orchard rows, where dense tree canopies can block GPS signals. Additionally, in environments with unknown obstacles, such as tree branches in orchard rows, human pilots are frequently queried to provide avoidance maneuvers, which significantly increases their workload. The ability to navigate autonomously and safely in orchard scenes with weak GPS signals and obstacles presents several challenges and largely hinders the deployment of UAVs in orchard operations. 

\begin{figure}[t!]
    \centering
    \includegraphics[width=0.85\linewidth]{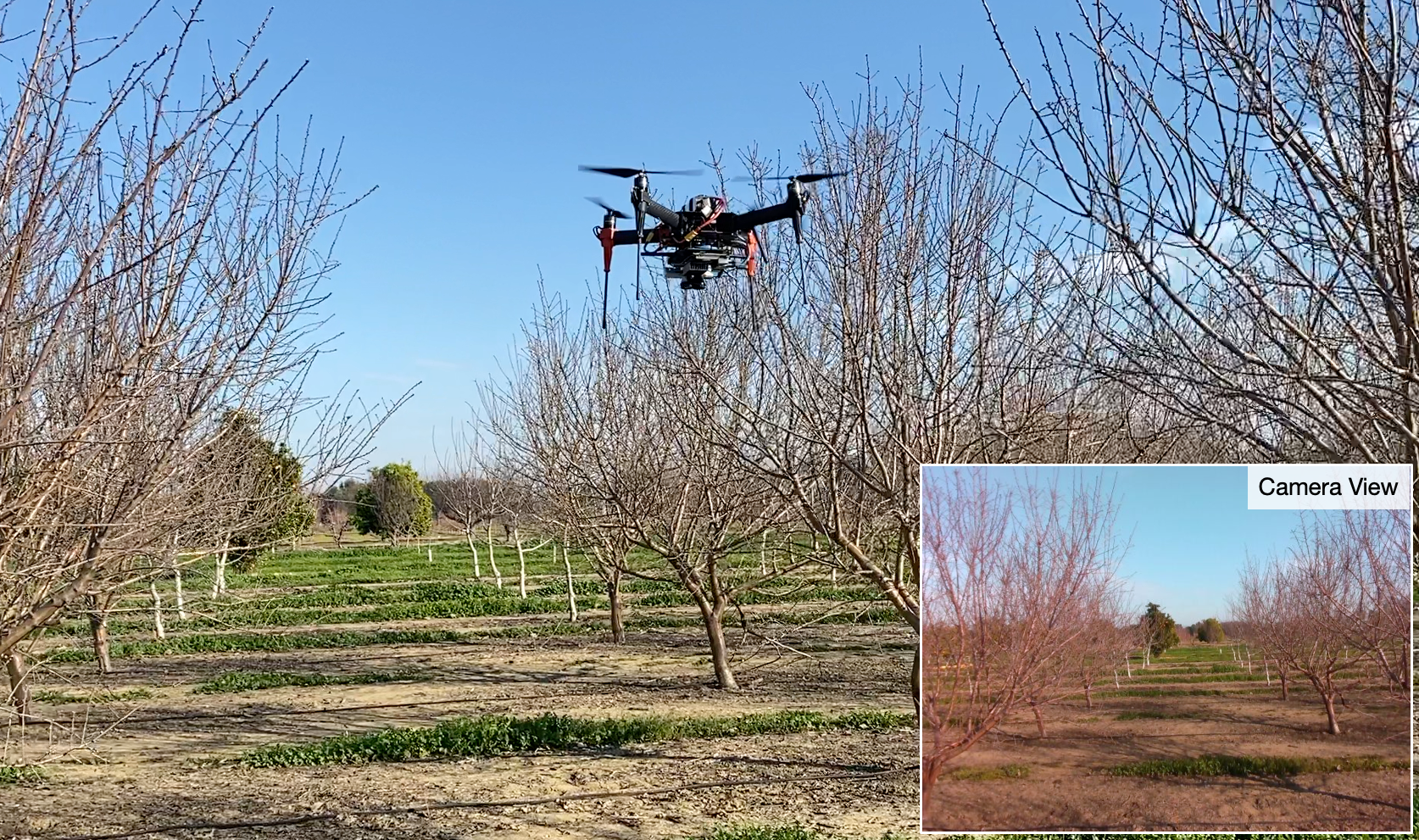}
    \caption{The drone is flying autonomously inside orchard rows without GPS, using its onboard camera to avoid tree branches along the way. The view of the onboard camera is provided.}
    \label{fig:drone_flight}
\end{figure}

When the GPS signal is attenuated, the UAV may rely on exteroceptive sensors to sense the environment and navigate. Advanced techniques to enable UAV autonomous operations without GPS include: 1) lidar-based, and 2) camera-based approaches. Typical lidar sensors are too large to fit on lightweight UAVs flying between the tree rows and can significantly reduce the flight time when size and endurance are considerations. It leaves the camera-based algorithm a better option for UAV autonomous navigation in orchards where compact size and longer battery life are pursued. Also, a camera typically costs less than a lidar sensor. 
 
In general robotics, a common approach to achieving autonomous navigation in unknown environments is to break the system into distinct modules, typically including perception, planning, and control. Each module is developed and optimized separately. For example, a global or local map of the environment can be constructed while the robot's state is estimated simultaneously. Within this map, a feasible path or trajectory is planned to reach the target point while avoiding obstacles. A tracking algorithm then controls the robot to follow the planned path. Although this modular design scheme has been explored extensively in the literature, the drawback is that the computation load can be intensive and it introduces discrepancies between different stages due to the hierarchical task decomposition. This raises the question of whether such a modular design structure is necessary. Humans, for example, can fly drones relying solely on visual inputs (e.g., first-person-view camera). They directly map the visual information to control commands and provide good reactive behaviors in unknown environments~\citep{pfeiffer2021human}. Inspired by this, we aim to develop a similar reactive policy that skips explicit maps and waypoint constructions. Such a visuomotor policy can be trained through modern learning-based approaches.

Reinforcement learning (RL) has shown promising results in fields like self-driving car and strategy games. However, RL is sample-inefficient, requiring substantial amounts of data to achieve good results. Designing an effective reward function is also challenging, and the trial-and-error nature of RL hinders its usage in safety-critical systems, where a single failure may lead to enormous losses. One way to address these shortcomings is to train a policy in simulation and transfer it to the real world during deployment. This transfer learning strategy reduces the costs and risks of training a policy by preventing system damage. However, inaccuracies in the simulation model (e.g., modeling of environmental changes can be difficult) may lead to unmatched performance in real-world applications, and minimizing this sim-to-real gap remains an ongoing challenge. In this work, we adopt an imitation learning (IL) framework to train a visuomotor policy. The robot learns a strategy by mimicking an expert's behavior from a small number of demonstrations. The aim is to leverage human knowledge to guide the learning process and achieve human-like behaviors while overcoming the limitations of the aforementioned methods.

In our previous work~\citep{peng2022imitation}, we proposed a variational autoencoder (VAE)-based controller trained via imitation learning and evaluated it in simulated riverine environments using Unreal Engine. The controller demonstrated competitive results compared to several vision-based navigation algorithms. However, the prior experiments were conducted entirely in simulation, where conditions such as obstacle structure, lighting, and vehicle dynamics were ideal and tightly controlled. In this paper, we extend that work by demonstrating, to the best of our knowledge, for the first time, the real-world deployment of a reactive imitation-based control policy for UAV navigation in complex orchard environments, as illustrated in Figure~\ref{fig:drone_flight}. The control policy is trained using an interactive imitation learning framework with image data and expert demonstrations collected exclusively in the field. This eliminates the common sim-to-real transfer issue present in many learning-based methods that rely on simulation pre-training. Real-world orchard environment introduces significant challenges compared to simulation, including unstructured obstacles such as protruding branches and foliage, varying illumination conditions, and unpredictable wind disturbances. The actual UAV system also brings uncertainties in dynamics and control, as well as sensor noises. Despite these challenges, our controller enables fully autonomous UAV flight within orchard rows without relying on pre-configured maps or waypoints. It achieves longer flying distances with reduced human intervention and demonstrates strong generalization capabilities in novel environments and under varying speeds. The contributions of this paper are summarized as follows:
\begin{enumerate}
    \item We propose a reactive strategy to enable UAV navigation in orchard rows using only visual inputs. A variational autoencoder-based controller is developed and trained through an intervention-based learning framework using exclusively real-world flight data. We demonstrate the practical deployment of this controller on a custom-built UAV platform under real orchard conditions. Compared to existing algorithms, our VAE-based controller requires significantly less human intervention and achieves autonomous navigation after just a few iterations of training.

    \item We evaluate the generalization performance of our controller in novel environments and under varying speeds. The results show that our VAE-based control policy performs well in unseen environments and is robust to speed changes. It enables the UAV to travel longer distances with less human assistance requirement, outperforming existing baseline algorithms.
\end{enumerate}

Our proposed VAE-based controller relies exclusively on RGB images, without utilizing depth measurements from stereo cameras or LiDAR sensors. By avoiding reliance on depth information, we significantly reduce system complexity, cost, weight, and power consumption. This RGB-only approach enables broader applicability and easier deployment of lightweight UAV systems in diverse agricultural environments, where depth sensors may struggle due to environmental complexities such as varying illumination conditions and complex canopy structures \citep{condotta2020evaluation}. In this study, our primary focus is the vision-based navigation of UAVs within orchard rows, where GPS signals are heavily obstructed by canopies. Row-to-row transitions are typically considered simpler due to stronger GPS signals and the absence of obstacles in headland areas, allowing a GPS-based planner to guide UAV transitions effectively, given the coordinate of the target rows. Integrating such a GPS-based planner for row transitions with our vision-based navigation system could be pursued in future work as part of a comprehensive orchard monitoring framework.

The rest of the paper is structured as follows: In Section~\ref{sec:related_work}, we review related work in the literature. Section~\ref{sec:methodology} introduces our proposed methodology and the data collection procedure in orchards. The results from flight experiments are presented in Section~\ref{sec:results} followed by our discussions in Section~\ref{sec:discussion}. Section~\ref{sec:conclusion} concludes the paper.

\section{Related Work}
\label{sec:related_work}
The use of small UAVs for orchard monitoring and management has grown increasingly popular in recent years. Common applications include crop analysis and yield estimation. For example, UAVs equipped with multispectral cameras have been used effectively for tree and plant counting~\citep{donmez2021computer}, as well as yield prediction~\citep{chen2022predicting}. UAV platforms have also been developed for phenotyping; \citet{lopez2019efficient} demonstrated this by employing an RGB-UAV system to analyze almond trees. Another important application is disease and pest monitoring. UAV imagery has proven valuable in detecting vine diseases~\citep{kerkech2018deep}, and recent work has utilized the YOLO algorithm on UAV images for pest detection in orchards~\citep{sorbelli2023yolo}. These studies highlight the crucial role UAVs play in maintaining crop health. In precision pesticide application, UAVs have been used to extract canopy volume information for targeted spraying~\citep{zhang2023design}. Additionally, UAVs have shown their utility in general orchard surveying and navigation. \citet{stefas2019vision}, for instance, demonstrated UAV navigation in apple orchards to achieve comprehensive yield coverage. UAVs have similarly been employed for surveying orchard landscapes. For a thorough review of UAV applications in orchard management, see \citet{zhang2021orchard}. These studies collectively underscore the growing impact of UAV technology in enhancing precision agriculture and improving orchard management practices.

Autonomous navigation in agricultural fields presents several challenges. State-of-the-art approaches typically divide the task into perception, planning, and control modules, each of which can be developed independently. In orchards, perception and state estimation are particularly critical, especially in GPS-denied environments. To address these challenges, \citet{kim2020path} introduced a CNN-based path detection method using camera images for autonomous travel. Additionally, 3D point cloud data can facilitate localization within orchard rows~\citep{zhang20133d}. Simultaneous localization and mapping (SLAM) is another widely used approach to simultaneously localize the robot and build environmental maps. For example, \citet{chen20213d} presented an eye-in-hand stereo vision and SLAM system capable of high-resolution global mapping for large-scale orchard picking tasks. These techniques provide essential inputs for subsequent trajectory planning. Once the environment is mapped, collision-free trajectories can be generated using various planning methods. Following this, tracking algorithms enable robots to execute planned paths accurately. Within this modular framework, \citet{jiang2024navigation} developed a navigation system for autonomous spraying robots in orchards, while \citet{pan2024novel} introduced an advanced perception and semantic mapping approach to enhance robot automation in unstructured orchard environments. Moreover, \citet{raikwar2022navigation} created a model-based navigation and control system tailored for ground robots operating in GPS-deprived orchards. Additionally, \citet{stefas2019vision} demonstrated a vision-based UAV system in apple orchards, effectively highlighting capabilities in autonomous navigation, obstacle avoidance, and comprehensive yield coverage.

Although the modular design paradigm has achieved success, the separation of perception and control can lead to unexpected behaviors. \citet{pfeiffer2021human} noted a discrepancy between trajectory planning and deployment, and the requirement to build a 3D map, search for collision-free paths, and solve constrained optimal control problems adds extra cost and delay to real-time systems. Conversely, human pilots are able to navigate robots in complex, unknown environments using only a camera stream, providing excellent reactive behaviors and generalizable skills across diverse scenarios~\citep[e.g.,][]{ross2013learning, loquercio2018dronet}. Inspired by this, a similar visuomotor policy, which directly maps states to actions, can be designed. Such policies can be represented by deep neural networks and trained end-to-end using various learning-based approaches.

Reinforcement learning (RL) seeks to find the optimal policy by maximizing a reward function. RL has been applied successfully in a wide range of applications. For example, \citet{song2021autonomous} applied proximal policy optimization (PPO) to train a drone to fly through race tracks at high speed. \citet{wurman2022outracing} used a soft actor-critic (SAC) algorithm to achieve super-human performance in autonomous car racing, and \citet{smith2022walk} applied deep reinforcement learning to quadruped locomotion in the real world. Despite these successes, RL suffers from poor sample efficiency, often requiring large datasets and slow convergence. Its trial-and-error nature poses risks for safety-critical systems, where mistakes can lead to significant losses.

Various transfer learning techniques have been explored to train and deploy models across different domains.  \citet{loquercio2019deep} successfully transferred a racing drone policy trained in simulation using domain randomization to a physical platform without fine-tuning. Domain adaptation helps bridge the simulation-to-reality gap, as seen in \citet{kang2019generalization}, who adapted models trained in simulation with real-world data for better generalization. However, transferring a simulation-trained model to the real world remains challenging due to the need for accurate modeling of physical dynamics and environmental factors, requiring extensive engineering effort.

Imitation learning (IL) offers an alternative by learning a policy from expert demonstrations, avoiding the need for a handcrafted reward function. Expert supervision accelerates the learning process and makes IL more data-efficient. This approach has proven effective in autonomous driving tasks. UAV navigation in GPS-denied environments has also been explored through IL. \citet{ross2013learning} trained drones to navigate low-altitude forests and avoid trees using data aggregation. \citet{giusti2015machine} trained a UAV to traverse forest trails using the monocular visual perception of trees and foliage. Additionally, \citet{gandhi2017learning} taught a quadrotor to avoid obstacles by learning from crash data. In urban environments, \citet{loquercio2018dronet} trained a UAV navigation policy using data from cars and bicycles, demonstrating strong generalization across both indoor and outdoor settings. IL has also been applied to achieve agile drone flight in dynamic environments \citep{kaufmann2018deep}.

Recent research has shown that using privileged information can enhance performance in traditional deep learning methods. In this approach, a teacher policy with access to privileged information is trained first, followed by a student policy that imitates the teacher without access to the privileged information. The student policy is then deployed in the real world. This method has proven successful in various applications, including vision-based urban driving~\citep{chen2020learning}, high-speed drone flight~\citep{loquercio2021learning}, and quadruped locomotion in challenging environments~\citep{Lee_2020}. However, selecting appropriate privileged information can be vague, and performance may not be guaranteed without training in high-fidelity simulators, a challenge particularly relevant to agriculture, where high-quality simulators are scarce.

\section{Materials and Methods}
\label{sec:methodology}
In this section, we first introduce our custom-built UAV platform. Next, we present the intervention-based imitation learning framework and discuss its advantages. We then describe the neural network structure of our vision-based policy, followed by the architecture of the entire system. Finally, we outline the data collection procedure conducted in the field.

\subsection{UAV Platform}
\label{sec:uav_platform}
\begin{figure}[t!]
    \centering
    \includegraphics[width=0.5\linewidth]
    {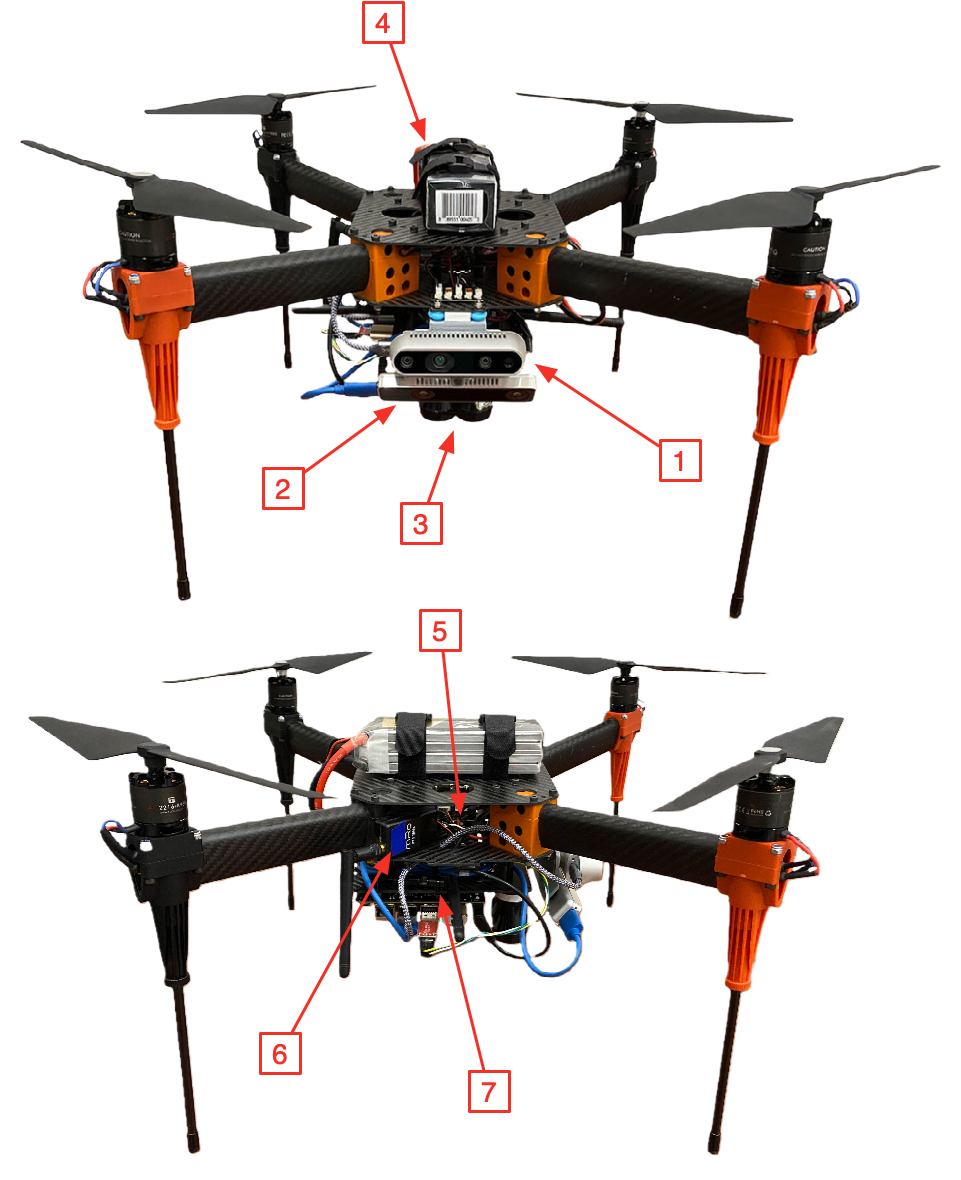}
    \caption{The custom-built quadcopter platform equipped with 1) Intel  RealSense D435i camera; 2) Intel RealSense T265 camera; 3) Lightware SF11/C laser rangefinder; 4) 4s 5200mhA LiPo battery; 5) Pixracer R15 flight controller; 6) mRo Ski Telemetry radio; and 7) Nvidia Jetson Xavier NX.}
    \label{fig:quadcopter}
\end{figure}

We developed a custom quadcopter platform for this research. The main frame has a wheelbase of 450 mm. A RealSense D435i camera (Intel Corporation, Santa Clara, California, USA) is mounted forward and streams RGB images with a field of view of 70$^{\circ}$. Visual odometry is provided by a RealSense T265 tracking camera (Intel Corporation, Santa Clara, California, USA). We found that the altitude measurement from the tracking camera was not accurate enough. Therefore, an SF11/C laser rangefinder Light(LightWare LiDAR, Boulder, Colorado, USA) is integrated to measure the height above ground. We chose the PixRacer R15 (mRobotics, Chula Vista, California, USA) as the flight unit, with the PX4 Autopilot stack as the low-level controller. The onboard computational unit is a Nvidia Jetson Xavier NX board (Nvidia, Santa Clara, California, USA), and an external solid-state drive is mounted to store the data. We also added a radio to receive real-time telemetry status from the drone. A 4S LiPo battery is used to power the entire system and can support a flight time of 13 minutes with a take-off weight of 1.8 kg. Pictures of the platform can be seen in Figure~\ref{fig:quadcopter}.

\subsection{Intervention-based Imitation Learning}
\label{sec:imitation_learning}

In this work, we use imitation learning to train a vision-based policy. As discussed above, imitation learning offers several advantages over reinforcement learning, such as reduced sampling complexity and eliminating the need for a hand-crafted reward function. A common strategy in imitation learning is to learn an agent policy $\pi_{\theta}$, parameterized by $\theta$, that mimics the expert policy $\pi_{E}$. The optimal policy $\pi_{\theta}^*$ is found by minimizing a similarity measurement $\mathbb{D}$ between two policies, as shown in \eqref{eq:imitation_learning}:
\begin{equation}
    \pi_{\theta}^* = \argmin_{\theta} \mathbb{D}(\pi_{\theta}, \pi_{E})
    \label{eq:imitation_learning}
\end{equation}

The simplest approach in IL formulates the problem as a supervised learning task,  with the objective of matching the learned policy $\pi_{\theta}$ to the expert policy $\pi_E$ based on a dataset of expert demonstrations $\mathcal{D}$ — a method known as behavior cloning. However, behavior cloning often fails in practice because the distribution of states encountered during the execution of learned policy $\pi_{\theta}$ may not match the distribution in the demonstrated dataset. To overcome this limitation, \citet{ross2010efficient} proposed a data aggregation (DAgger) method, which uses a queryable expert to provide new demonstrations during the learning process. Despite its improvements, the DAgger algorithm has two major drawbacks. First, the querying step is very inefficient, as the expert must explicitly label all new data points. Second, during the agent policy rollout, the expert lacks direct feedback from the robot and is, therefore, unaware of how their corrections affect future states.

\makeatletter  
\def\BState{\State\hskip-\ALG@thistlm}  
\makeatother 
\begin{algorithm}[t!]
\caption{Intervention-based Imitation Learning}\label{euclid} 
    \begin{algorithmic}
    \State \textbf{Input:} $\text{maximum iteration number}~i_{max}$
    \State \textbf{Output:} $\text{optimal agent policy}~\pi_{\theta}^*$
    \State \textbf{Initialize:} $i = 0, \mathcal{D} = []$
    \If {$i = 0$} 
        \State append all human demonstrations to $\mathcal{D}$
        \State train agent policy $\pi_{\theta_0}$ on $\mathcal{D}$
    \Else
        \For {$i=1:i_{max}$}
            \State roll out agent policy $\pi_{\theta_{i-1}}$ with human interventions
            \State get a series of rollout trajectories $\mathcal{T}_j$
            \For {each $(s_{t}, a_{t}) \in \mathcal{T}_j$}
                \If {$(s_t, a_t) \in \pi_{E}$} 
                    \State {append $(s_t, a_t)$ to $\mathcal{D}$}
                \EndIf
            \EndFor
            \State re-train agent policy $\pi_{\theta_i}$ on current $\mathcal{D}$
        \EndFor
    \EndIf
    \State $\pi_{\theta}^* \gets \pi_{\theta_i}$
    \State \textbf{return} $\pi_{\theta}^*$  
    \end{algorithmic}  
\label{algo:imtation_learning}
\end{algorithm}  

In this work, we utilize an intervention-based DAgger approach to train the agent policy. Instead of querying an expert offline, we query the expert online while the drone executes the agent policy in real time. During the rollout, the human pilot can intervene and take control if they believe the agent's actions might lead to failure or an unsafe state. Once the drone reaches a safe state, as defined by the pilot, control is returned to the agent policy. Only the demonstrations where the pilot takes control are appended to the training dataset $\mathcal{D}$, and the policy is retrained after each iteration, following the standard DAgger process. Since the human expert provides feedback intermittently during execution, the querying procedure is significantly simplified. Moreover, the expert maintains uninterrupted control during demonstrations and receives real-time feedback from the drone. This interaction is more natural in real-world settings and enhances the quality of the collected data. A pseudo-code of our intervention-based learning algorithm is provided in Algorithm~\ref{algo:imtation_learning}.

\subsection{Variational AutoEncoder (VAE)-based Controller}
\label{network_architecture}

Our vision-based navigation policy consists of two components: a variational autoencoder network and a policy network, together referred to as the VAE-based controller. The details of each component are provided in the following sections.

\subsubsection{Variational AutoEncoder Network}

We incorporate a variational autoencoder network~\citep{kingma2013auto} into our navigation policy. As demonstrated in previous work~\citep{peng2022imitation}, the VAE effectively extracts useful information from high-dimensional data, reducing it into a lower-dimensional space, which improves both training efficiency and generalization in vision-based navigation tasks. A VAE typically includes two components: an encoder and a decoder. The encoder compresses image data from a high-dimensional space into a lower-dimensional space and generates latent variables that capture dominant aspects of the original images (e.g., position, scale, rotation, lighting). These latent variables are then passed to the decoder, which reconstructs the image. 

To train the VAE, we minimize the loss function in \eqref{eq:VAE_loss}. The first term represents the reconstruction error between the original image $x$ and the reconstructed image $\hat{x}$, computed by the decoder $p_{\varphi}(\hat{x}|z)$, where $z$ are the latent variables and $\varphi$ represents the decoder network parameters. The second term is the Kullback–Leibler (KL) divergence between the latent variables distribution $q_{\psi}(z|x)$ extracted by the encoder network (parameterized by $\psi$), and Gaussian distributions $p(z)$. The reparameterization trick is applied to ensure gradients can be properly backpropagated. We use a weight factor $\beta$, set to 3 in this paper, to balance the reconstruction error and KL divergence. 
\begin{equation}
    \mathcal{L}_{VAE}(\varphi, \psi) =\Vert x - p_{\varphi}(\hat{x}|z) \Vert^2 - \beta \cdot D_{KL}(q_{\psi}(z|x),p(z))
    \label{eq:VAE_loss}
\end{equation}

\begin{figure}[t!]
    \centering
    \includegraphics[width=1\linewidth]{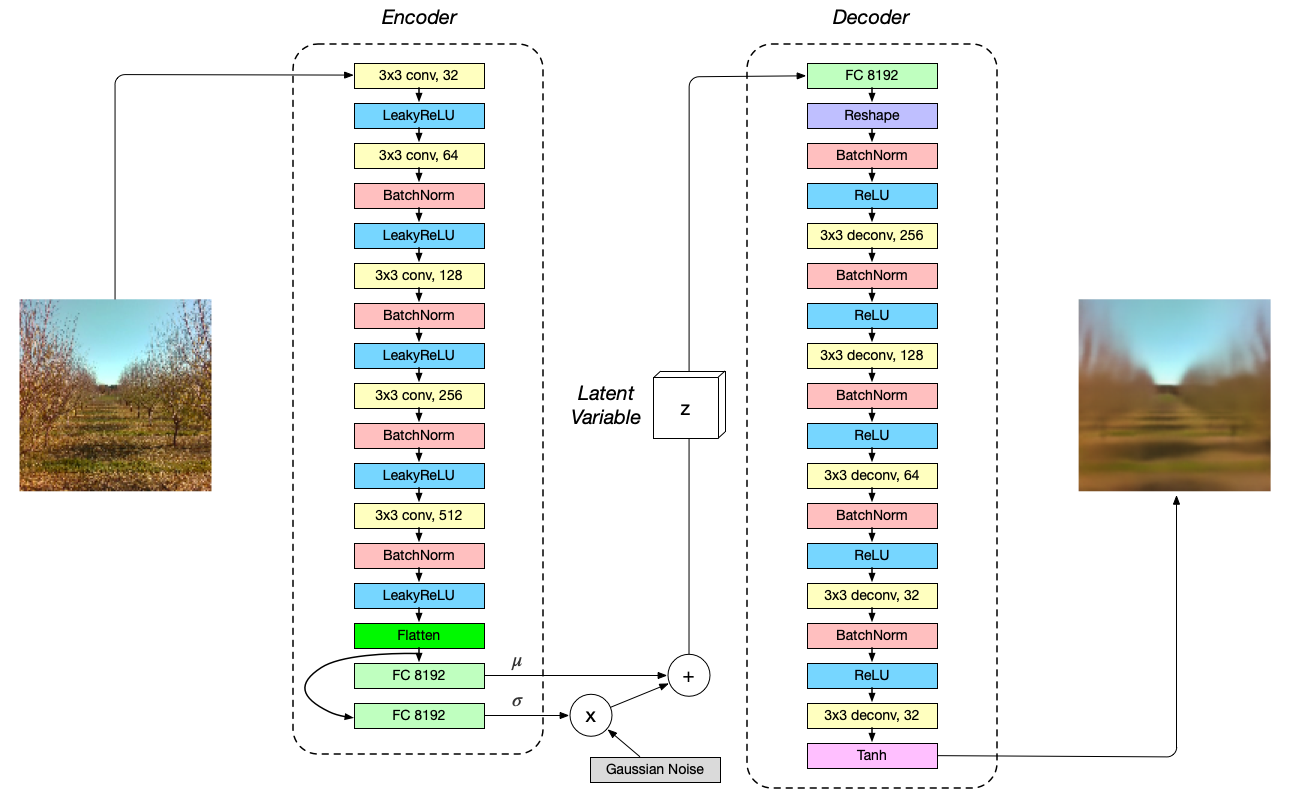}
    \caption{The structure of the variational autoencoder network.}
    \label{fig:vae_structure}
\end{figure}

The structure of our VAE is illustrated in Figure~\ref{fig:vae_structure}. The encoder consists of five convolutional layers with LeakyReLU activations, while the decoder has five transpose convolutional layers with ReLU activations. The encoder takes an input image of size $128 \times 128 \times 3$ and outputs latent variables of dimension 256. The decoder reconstructs the image to the same size. Figure~\ref{fig:vae_structure} also shows an example of an original image and its reconstruction, illustrating the VAE's performance. During deployment, we use only the trained encoder to compute the latent variables (with frozen weights), which are then fed into the policy network, described in the next section.

\subsubsection{Policy Network}
Our policy network is based on a multilayer perceptron (MLP) with two hidden layers, each consisting of 256 units with ReLU activations. The output layer uses a hyperbolic tangent activation function to generate the control command within the range $[-1, 1]$. The inputs to the policy network are the latent vector extracted by the VAE encoder and the current drone state, which includes attitude and velocity information from the state estimator. We train the network by minimizing the mean square error (MSE) between the output of the agent policy $\pi_{\theta}(a|s)$ and the human expert policy output $\pi_{E}(a|s)$, as shown in~\eqref{eq:mlp_loss}, using the dataset of expert demonstrations $\mathcal{D}$.
\begin{equation}
    \mathcal{L}_{policy}(\theta) = \Vert \pi_{\theta}(a|s) - \pi_{E}(a|s) \Vert^2,~s\in \mathcal{D}
    \label{eq:mlp_loss}
\end{equation}

\subsection{System Architecture}
\begin{figure}[t!]
    \centering
    \includegraphics[width=1.0\linewidth]
    {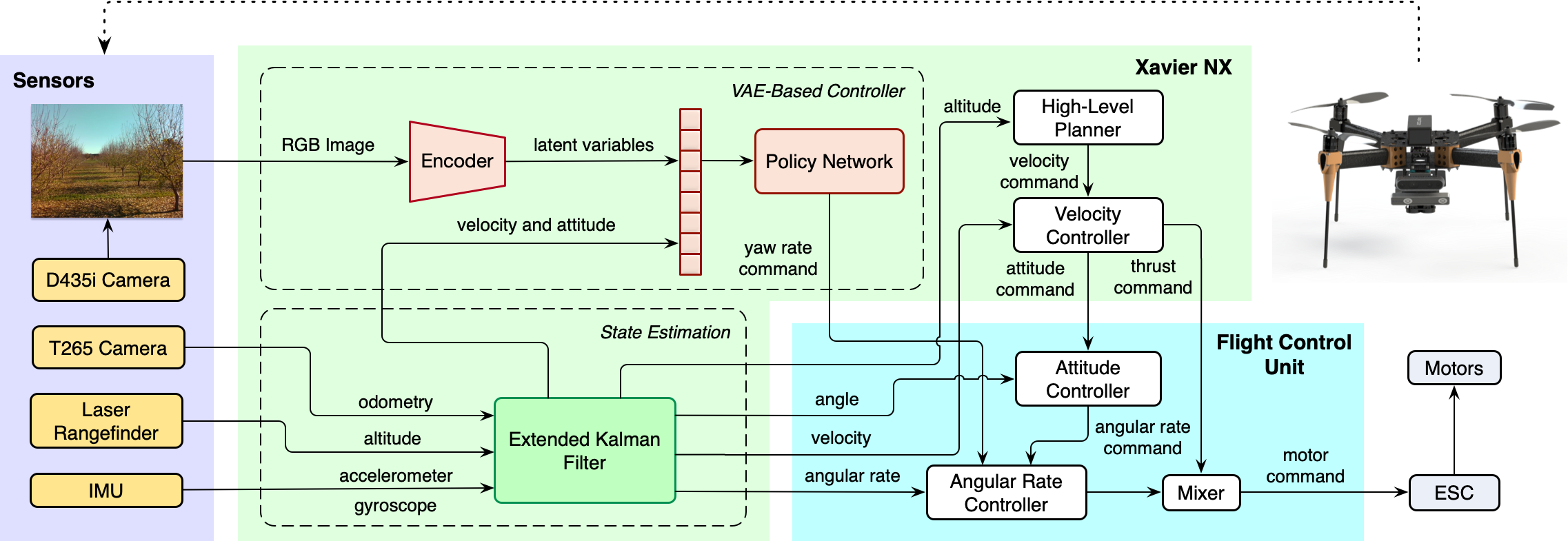}
    \caption{System architecture diagram during training and deployment.}
    \label{fig:system_architecture}
\end{figure}

We present the system architecture in Figure~\ref{fig:system_architecture}. The RGB images are streamed from the D435i camera at 30 Hz. We chose the T265 camera for its internal visual inertial-odometry (VIO) pipeline, though other VIO methods could also be used at the cost of increased computational load on the onboard computer. We fuse the visual odometry from the T265 camera, altitude measurement from the laser rangefinder, and IMU measurements using an Extended Kalman Filter (EKF). The EKF estimates the altitude, orientation angles, linear velocities, and angular rates of the drone. A high-level planner uses the estimated altitude as input to generate velocity commands that help the drone maintain a constant height and forward speed. Velocity tracking is achieved by a PID-based velocity controller, which generates target thrust and attitude commands. These attitude commands are passed to an attitude controller, which outputs target angular rates for the angular rate controller. The angular rate controller then computes the required torques to achieve the commanded angular rates. This process follows a standard cascaded PID control architecture. Concurrently, our VAE-based controller generates yaw rate commands at 30 Hz for orchard navigation, which are directly fed into the angular rate controller. The mixer onboard the UAV converts these torque and thrust commands into individual motor signals. Specifically, it determines the appropriate combination of motor outputs to achieve the desired total thrust and rotational moments. These commands are sent to the electronic speed controllers (ESCs), which actuate the individual motors. State estimation, neural network inference, and high-level control are executed on the Jetson Xavier NX, while low-level flight control including the cascaded PID controllers and mixer is managed by the PX4 flight stack, which generates the final motor commands. The PX4 communicates with the Xavier NX through the Robot Operating System (ROS) interface. The onboard computer is also connected to a laptop via Wi-Fi, allowing users to start scripts and monitor the system's status. Human pilots can teleoperate the drone using a remote transmitter to provide demonstrations for training or take control in emergencies. 

\subsection{Data Collection and Training}
\label{sec::data_collection_in_real_world}

\begin{figure}[t!]
    \centering
    \includegraphics[width=0.95\linewidth]{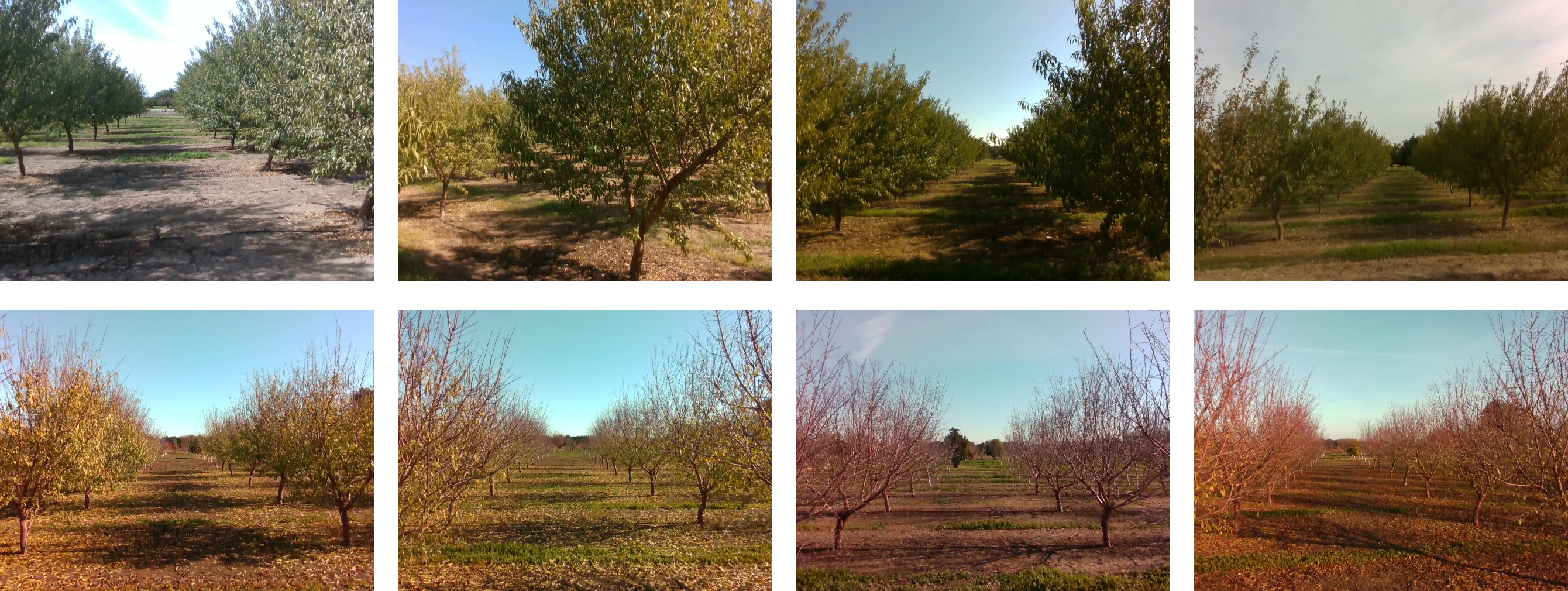}
    \caption{Sample image data collected in a variety of seasons, weather conditions, and times of day in the mixed-species orchard.}
    \label{fig:field_images}
\end{figure}

Our training data were collected in an orchard near Davis, California, USA, during 2022-2023. This orchard is a research facility containing a mix of almond, walnut, plum, pistachio, and peach trees, with an average tree age of three years. We refer to this as the mixed-species orchard throughout the paper. The trees are planted in rows spaced 20 ft (6.10 m) apart and with a tree-to-tree distance of 15 ft (4.57m). Each row is approximately 240 ft (73.15 m) in length. Specifically, we selected six rows from this orchard for image data collection to train our VAE network. Each selected row includes mixed tree species, providing a visually diverse environment. To further enhance the variability of our dataset, images were collected across different seasons (summer, autumn, winter), weather conditions (sunny, cloudy), and times of day (morning, noon, afternoon). Samples of the collected images are shown in Figure~\ref{fig:field_images}. In total, the VAE training dataset contains 110,551 images with a resolution of 640 x 480 pixels.

We applied standard data augmentation techniques, including adjustments to brightness, contrast, saturation, and sharpness, as well as random horizontal flipping. All training was performed on a lab computer equipped with an AMD Ryzen 9 3900X CPU, 64 GB of memory, and an Nvidia GeForce RTX 2080Ti GPU, using the ADAM optimizer. Once the VAE network was trained, we froze the encoder's weights and used it to generate latent variables for training the policy network. We have made our dataset publicly available at: \url{https://drive.google.com/drive/folders/1tZiZu2b680ZqZX4IVkYo1QmVzZMHSpVP?usp=drive_link}.

The policy network was trained using the procedure outlined in Algorithm~\ref{algo:imtation_learning} from Section~\ref{sec:imitation_learning} and with the ADAM optimizer. Human demonstrations were collected from the same six orchard rows used previously for VAE data collection. Initial demonstrations were gathered in late summer and autumn, while subsequent iterations of imitation learning took place during autumn and winter seasons. Given the multi-model nature of human behavior, we observed that allowing the human pilot full control over the drone led to inconsistent training results. For instance, the pilot might fly at a high altitude to avoid all trees or command the drone to fly very slowly, resulting in too conservative behavior. This expanded the state space and made training difficult. Therefore, in our approach, we only allowed the pilot to provide yaw rate correction, while a high-level controller was implemented to keep the drone flying at a constant height above the ground and at a constant forward speed in the body-fixed frame. It's worth mentioning that navigating within orchard rows is a challenging task due to unreliable GPS signals, unstructured obstacles (e.g., tree branches), unpredictable environmental disturbances (e.g., wind gusts), and the imperfect balance of the drone. As a result, the drone does not always fly in a straight line and may exhibit lateral drift.  Continuous yaw corrections are required to keep the drone in safe regions, making this task a suitable application for our vision-based algorithm. During the initial human demonstration, we collected 23,557 demonstration instances. After the first iteration, an additional 5,193 demonstrations were gathered. Following the second and third iterations, we collected 2,507 and 630 demonstration instances, respectively. Overall, our training dataset comprised 31,887 expert demonstrations, equivalent to approximately 35 minutes of flight data. Compared to the 110,551 images (2 hours of data collection) used for training the VAE, the expert demonstrations constitute about $28.8\%$ of the total image data collected. This distribution highlights the effectiveness of our approach, where the VAE benefits from a larger and more diverse image dataset to robustly encode visual features, while the control policy efficiently learns from a smaller set of expert demonstrations within the latent space.

During the human demonstration data collection, we set the flight height between 1.6 and 2.0 meters to avoid the ground effect and to ensure obstacles were visible in the camera's view. The drone was commanded to fly forward at a speed of 0.6 m/s, allowing the pilot enough time to provide effective demonstrations. To prevent large state estimation errors, we limited the maximum yaw angular speed to 45 deg/s, as the camera used in these experiments tends to lose tracking under rapid yaw movements. Ideally, the human pilot would share the same view as the agent via a real-time video feed from the onboard camera. However, achieving this in practice is challenging due to technical constraints. Real-time streaming is often unstable due to the limited range of Wi-Fi signals, further degraded by the trees in orchards. Moreover, wireless transmission introduces additional time delays that are difficult to quantify. As a result, we provided the human pilot with full state information about the environment. Despite this limitation, we found out that the agent was still able to learn an effective policy. The results are presented in the next section.

\section{Experimental Results}
\label{sec:results}

In this section, we present the results of our vision-based policy during training and evaluate its performance in unseen environments. We also compare its performance with a set of baseline algorithms from the literature and show the results.

\subsection{Qualitative Results}
\label{sec:qualitative_results}
\begin{figure}[t!]
    \centering
    \includegraphics[width=0.7\linewidth]
    {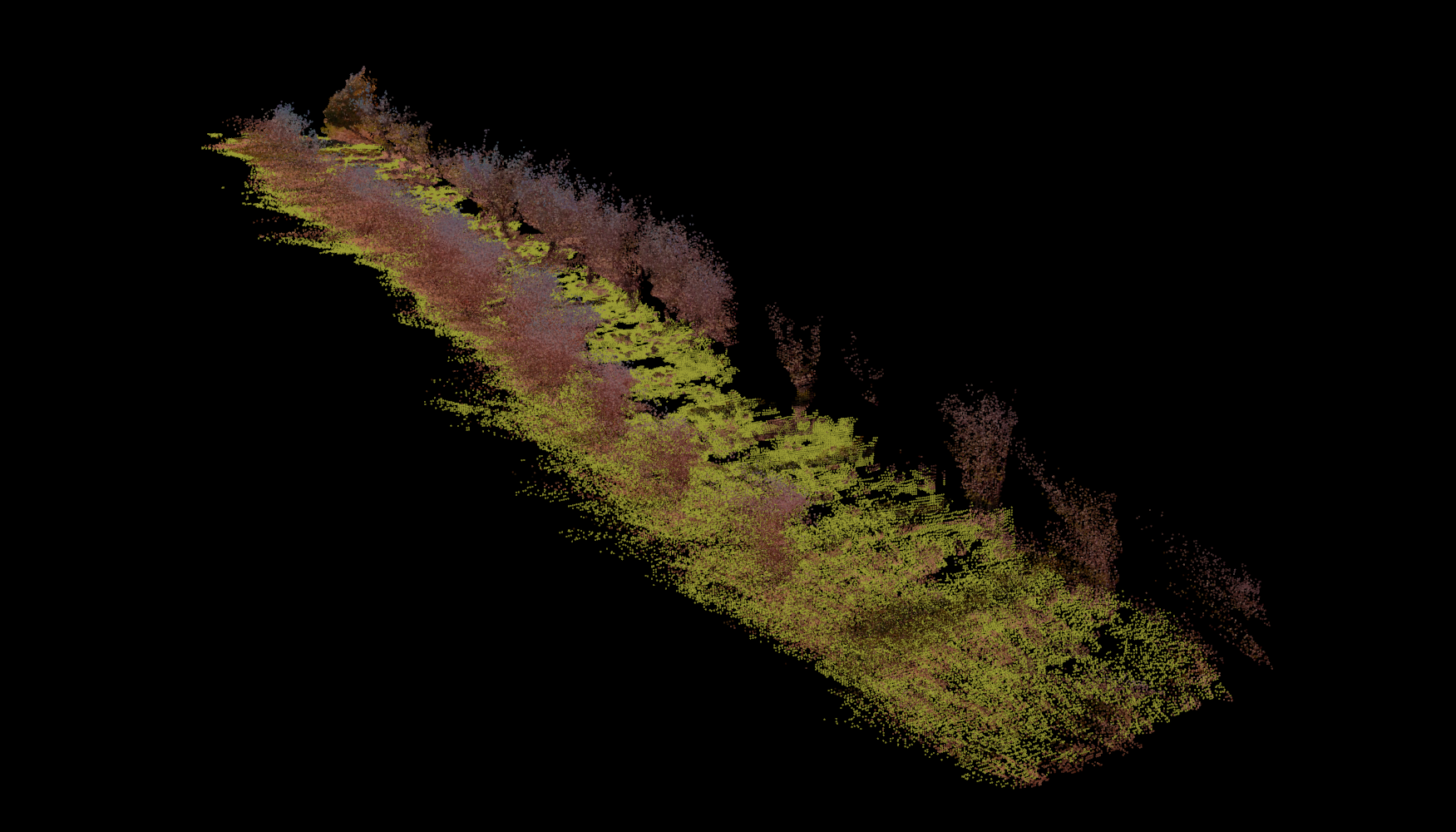}
    \caption{A filtered point cloud generated from the experiment data. The yellow-green points represent the ground and the brownish-red points represent the trees.}
    \label{fig:point_cloud}
\end{figure}

\begin{figure}[t!]
    \centering
    \includegraphics[width=0.9\linewidth]
    {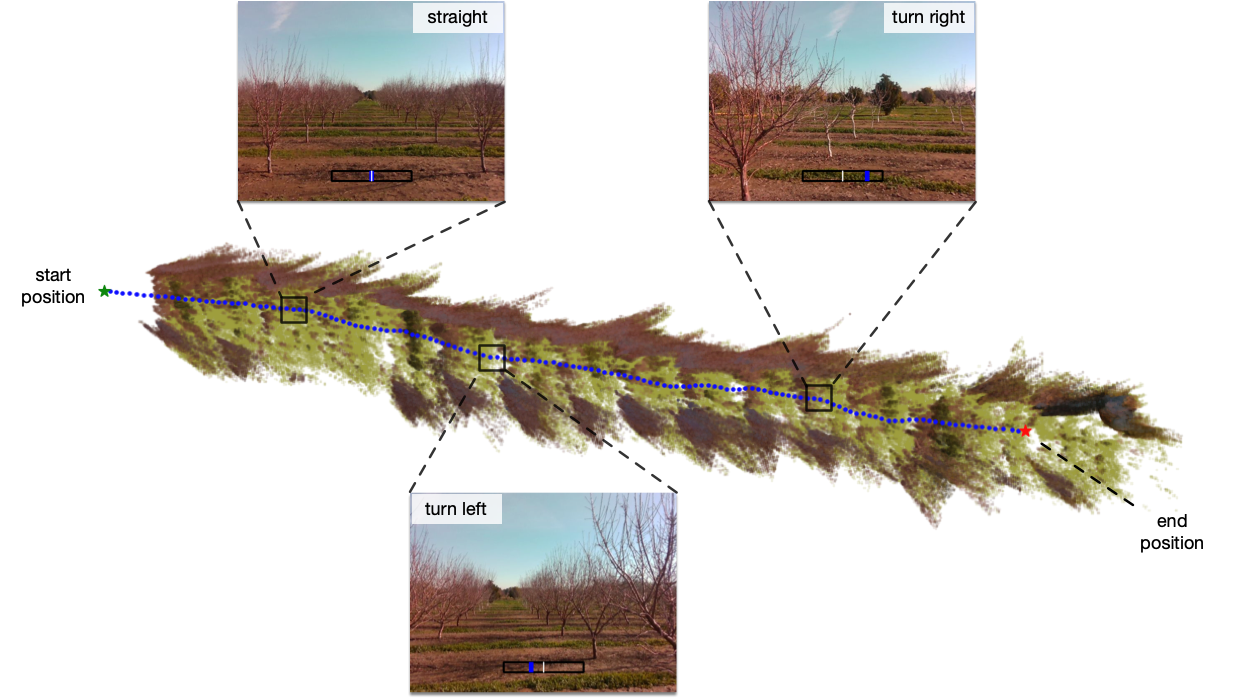}
    \caption{Top-down view of the trajectory performed by the proposed VAE-based controller after three iterations of training. The views from the onboard camera at different locations are provided.}
    \label{fig:trajectory_three_iteration}
\end{figure}

We first show the qualitative results of our proposed VAE-based controller. The drone was flown in the selected six rows mentioned earlier, and our VAE-based controller was trained through the intervention-based learning method described in Section~\ref{sec:imitation_learning}. To visualize the evolution of the agent policy's performance across training iterations, we used an open-source SLAM library~\citep{labbe2019rtab} to plot the drone's trajectory in a reconstructed 3D map offline. Since the D435i camera has a stereo setup capable of generating depth images, we fused the depth and RGB images with the drone's localization results into the SLAM library, creating a point cloud map of the environment. The raw point cloud was filtered to remove outliers and irrelevant points, such as the sky. The ground was identified using the RANSAC algorithm and re-colored for better visualization. A sample image of the filtered point cloud can be seen in Figure~\ref{fig:point_cloud}. Figure~\ref{fig:trajectory_one_iteration} shows a top-down view of the drone's trajectory within the filtered point cloud map after one iteration of training. The trajectory controlled by the agent policy is plotted in blue, while human intervention is shown in red. As observed, the agent policy learned some basic skills after one iteration but still required human assistance at times to avoid obstacles and navigate in the field. Based on the experiment, we found that performance improved significantly after three iterations of training. Figure~\ref{fig:trajectory_three_iteration} shows a sample flight trajectory in the map after three iterations of training, where the agent successfully learned to avoid obstacles and achieved fully autonomous flight without human intervention. For a supplementary video, see: \url{https://www.youtube.com/watch?v=VPlTaJbdo7U}

\begin{figure}[t!]
    \centering
    \includegraphics[width=0.9\linewidth]
    {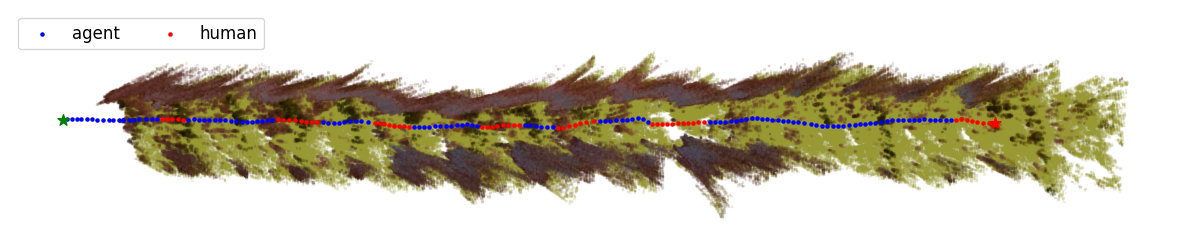}
    \caption{Top-down view of the trajectory performed by the proposed VAE-based controller after one iteration of training.}
    \label{fig:trajectory_one_iteration}
\end{figure}

\begin{figure}[t!]
    \centering
    \includegraphics[width=0.9\linewidth]
    {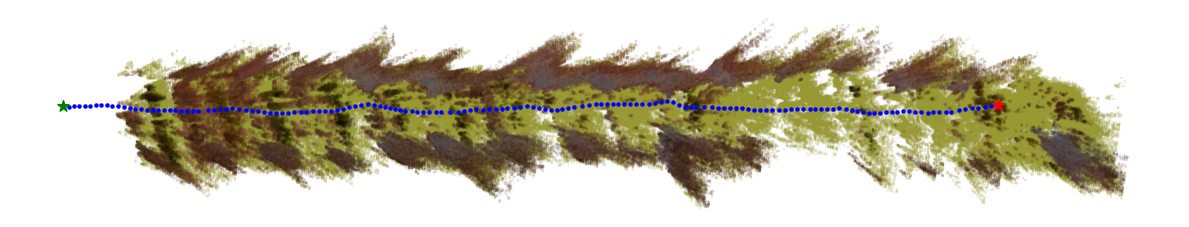}
    \caption{Top-down view of the trajectory executed by the VAE-based controller in a novel mixed-species orchard row}.
    \label{fig:trajectory_testing}
\end{figure}

A policy trained to perform well in training environments may not necessarily generalize to unseen environments. To evaluate the generalization performance of our vision-based policy in novel environments, we selected two additional rows from the mixed-species orchard that the agent had not encountered during training. This evaluation was conducted in winter 2023 to assess how effectively the trained policy adapts to previously unseen environments. We deployed our VAE-based controller, trained after three iterations, in these new rows and observed that the drone flew successfully with minimal human intervention. An example of a fully autonomous trajectory performed by our VAE-based controller in one of these novel rows is shown in Figure~\ref{fig:trajectory_testing}.

\begin{figure}[t!]
    \centering
    \includegraphics[width=1\linewidth]
    {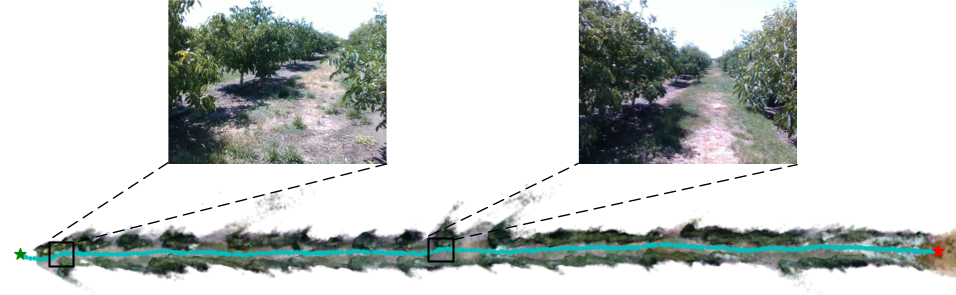}
    \caption{Top-down view of the trajectory executed by the VAE-based controller in a novel walnut orchard row. The controller successfully completed the navigation without human intervention.}
    \label{fig:trajectory_testing_walnut}
\end{figure}

\begin{figure}[t!]
    \centering
    \includegraphics[width=1\linewidth]
    {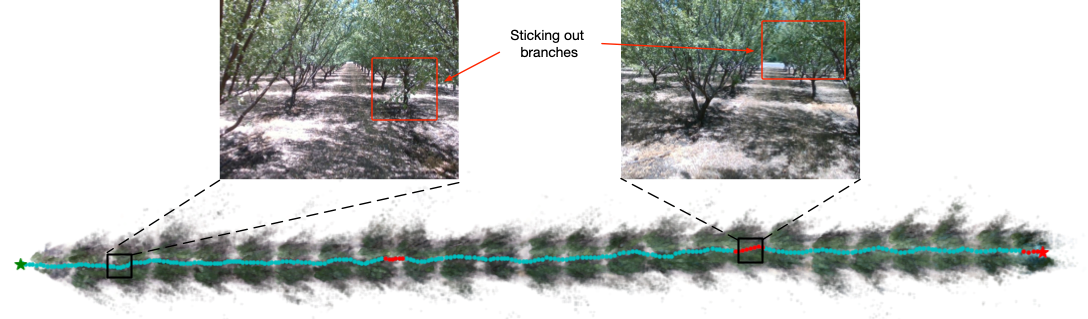}
    \caption{Top-down view of the trajectory executed by the VAE-based controller in a novel almond orchard row. Segments navigated autonomously by the agent are shown in cyan, while those requiring human intervention are shown in red. Two locations with protruding branches are indicated: at the first location, the agent successfully navigated through without assistance, while human intervention was required at the second location.}
    \label{fig:trajectory_testing_almond}
\end{figure}

To further evaluate the generalization performance of our method under diverse conditions, we conducted additional experiments at new almond and walnut orchards during late spring 2025. These orchards were previously unseen by the UAV. The walnut orchard consists of rows spaced 20 ft (6.10 m) apart, with a tree-to-tree spacing of 20 ft (6.10 m) and row lengths of approximately 500 ft (152.4 m). The almond orchard has narrower row spacing of 15 ft (4.57 m) apart, with the same tree-to-tree distance and row lengths of around 480 ft (146.3 m). During this season, foliage density was significantly higher, with closely spaced and protruding branches commonly observed. In these experiments, particularly in the almond orchard, the UAV was often required to take sharp turns to avoid protruding branches and dense foliage. Figure \ref{fig:trajectory_testing_walnut} illustrates the UAV trajectory executed by our VAE-based controller in a walnut row, along with representative images of the environment. The results demonstrate successful autonomous navigation throughout the test row. Figure \ref{fig:trajectory_testing_almond} presents the UAV trajectory in an almond row, accompanied by representative images. The results show effective navigation by our controller along most of the row, although human intervention was required at certain challenging locations to ensure safe obstacle avoidance. These segments are highlighted in red in Figure \ref{fig:trajectory_testing_almond}. Overall, these results indicate that our trained VAE-based controller generalizes well to diverse and challenging orchard environments.

\subsection{Quantitative Results}
\label{sec:quantitative_results}

To provide quantitative results and highlight the advantages of our VAE-based controller, we compared it with two baseline algorithms from the literature. 1) The first is a non-neural-network-based controller~\citep{ross2013learning}, in which the control policy is trained using human-selected visual features and computed through linear regression. We refer to this as \textit{baseline1}. 2) The second baseline is a neural-network-based controller~\citep{loquercio2018dronet} that employs a compact convolutional neural network (CNN) to infer control commands, referred to as \textit{baseline2}. Both controllers provide reactive behaviors similar to ours and have demonstrated effective performances in real-world scenarios. For a fair comparison, we modified both original implementations so that all controllers use input image size of $128 \times 128 \times 3$ and output yaw rate commands, as in our approach. All three controllers were trained using our intervention-based learning method with the same number of iterations from the human pilot. To minimize potential bias from human awareness, the sequence of controller execution was randomized during the experiment, ensuring that the pilot did not know which controller was being trained.

\begin{table}[b!]
\centering
\caption{T-Test results for human intervention rate with a significance level $\alpha=10\%$.}
\begin{NiceTabular}{lcccccc}
    \toprule
    & \multicolumn{3}{c}{\textbf{VAE-based (proposed) vs. Baseline1}} & \multicolumn{3}{c}{\textbf{VAE-based (proposed) vs. Baseline2}} \\
    \cmidrule[0.5pt](lr){2-4}\cmidrule[0.5pt](lr){5-7}
    \bf{T-Test} & \bf{iteration 1} & \bf{iteration 2} & \bf{iteration 3} & \bf{iteration 1} & \bf{iteration 2} & \bf{iteration 3} \\
    \midrule
    t-score & 1.3125 & 3.3187 & 3.7547 & 2.8887 & 3.1981 & 1.9527 \\
    p-value & 0.2058 & 0.00507 & 0.00274 & 0.00941 & 0.00644 & 0.07458 \\
    \bottomrule
\end{NiceTabular}
\label{tab:t_test}
\end{table}

\begin{figure}[t!]
    \centering
    \includegraphics[width=0.75\linewidth]{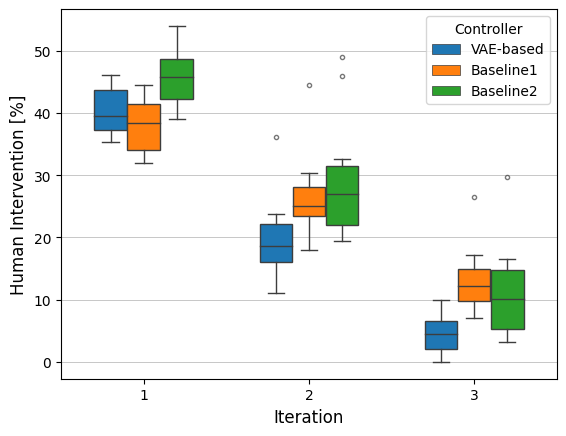}
    \caption{The box plot of human intervention rate against training iterations for all three controllers.}
    \label{fig:human_intervention_rate}
\end{figure}

We first calculated the human intervention rate during the training process for all three controllers, and the results are provided in Figure~\ref{fig:human_intervention_rate}. Each controller was tested in the six training orchard rows for three iterations, with demonstration data collected from the same human pilot. The box plot shows the mean and distribution of the human intervention rate, with a 95\% confidence interval at different iterations. As expected, the intervention rate decreases as the training progresses for all three controllers, demonstrating the effectiveness of our learning framework. Among the three controllers, we observe that the VAE-based controller maintains the lowest averaged intervention rate across all iterations, except the first iteration, where baseline1 shows a slightly lower intervention rate. After three iterations of training, the VAE-based controller reduces the average intervention rate to below 10\%, whereas the other controllers require more human assistance. To rigorously validate our results, we conducted a t-test to assess the significance of the differences in human intervention rates between our VAE-based controller and the baselines. The null hypothesis is that there is no significant difference in intervention rates between the controllers. The t-test results are presented in Table~\ref{tab:t_test}. With a significance level $\alpha = 10\%$, a p-value smaller than $\alpha$ means that we can reject the null hypothesis. Based on the calculated p-values and the mean values, we conclude that our controller outperforms both baseline controllers with significance.

\begin{figure}[t!]
    \centering
    \includegraphics[width=0.75\linewidth]{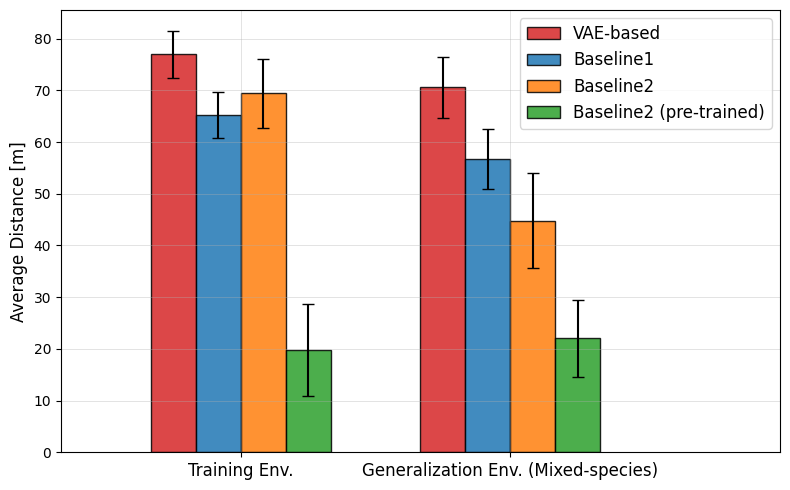}
    \caption{Average distance traveled before a failure for all controllers in training and generalization environments in the mixed species orchard. }
    \label{fig:average_distance}
\end{figure}

Next, we evaluated the flight performance of fully trained models (after three DAgger iterations) in both training and generalization environments. In this experiment, we included the pre-trained version of baseline2 with slight modification on its output, referred to as \textit{baseline2 (pre-trained)}. We used the average distance flow by the drone before a failure (i.e., human intervention) as the evaluation metric. The experiments were carried out in two selected rows from the training environments and two new rows from the mixed species orchard that had not been seen previously. Each controller was tested over ten flights in both training and generalization environments with varying weather conditions (sunny, cloudy, windy) and times of day (morning, afternoon). The flight data were recorded and the average distances before failure are computed and shown in Figure~\ref{fig:average_distance}. From the results, it can be seen that performance drops when the controllers are deployed in unseen environments. Nonetheless,  the VAE-based controller consistently outperforms baselines, achieving longer flights before requiring human intervention in both training and generalization environments. During flight experiments, we observed that our VAE-based controller could autonomously navigate the drone most of the time, while the baselines needed more human assistance. Interestingly, baseline2 did not perform well in our experiments, likely because it relies on line-like features that were not present in the orchard environment. A detailed discussion of why our VAE-based controller performs better is provided in Section~\ref{sec:discussion}. 

\begin{figure}[t!]
    \centering
    \includegraphics[width=0.75\linewidth]{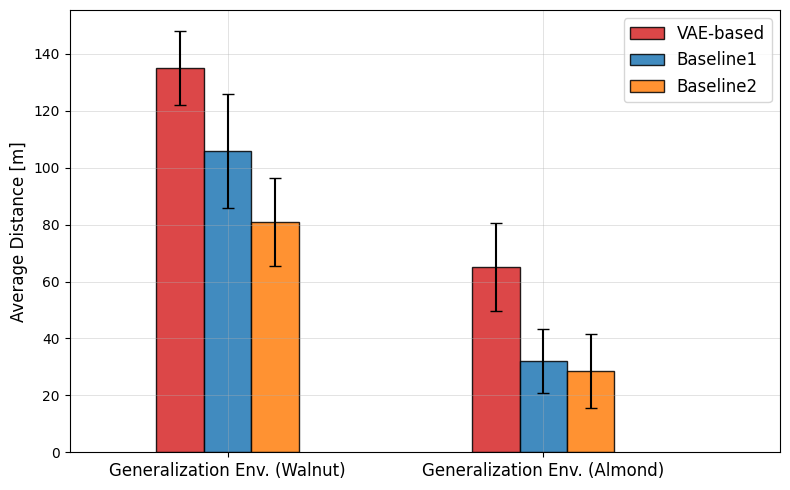}
    \caption{Average distance traveled before a failure for all controllers in the  walnut and almond generalization orchard environments}.
    \label{fig:average_distance_spring}
\end{figure}

Additional experiments were conducted in novel almond and walnut orchards during late spring 2025 to further evaluate the generalization performance of our controller. Descriptions of these two orchards are provided in the previous section. For each orchard, two rows were selected, and experiments were performed both in the morning and afternoon to capture a range of lighting conditions. Given that these rows are longer than those in the previously evaluated mixed-species orchard, three flights were conducted per row to ensure robust performance assessment. Considering the increased environmental complexity and previously poor performance of the baseline2 (pre-trained) controller, we only compared our VAE-based controller against baseline1 and baseline2 controllers in this experiment. We computed the average travel distance achieved by each controller in both environments, and the results are presented in Figure \ref{fig:average_distance_spring}. Overall, the walnut orchard is relatively easier to navigate, while the almond orchard posed more challenges due to its denser foliage and narrower row spacing. As a result, all three controllers showed reduced travel distances in the almond environment. Despite increasing difficulty, the VAE-based controller consistently outperformed the baseline methods, demonstrating strong robustness and generalization capabilities in complex and previously unseen orchard environments.

The processing time for each algorithm running on the Jetson Xavier NX was recorded and presented in Table~\ref{tab:process_time}. Baseline2 (pre-trained) was excluded from this comparison due to its similarity with baseline2. Additionally, all neural network components were recompiled and accelerated using TensorRT on our device, and the computer vision tasks in baseline1 also leveraged the onboard GPU for acceleration. From the results, it is evident that both neural network-based methods (baseline2 and ours) process images and generate inference outputs faster than the regression-based method in baseline1. This is because baseline1 requires explicit feature extraction, such as optical flow, from raw images, leading to a longer processing time. Baseline2 is slightly faster than our VAE-based controller due to its more compact network structure. Both neural network-based methods could run in real-time on the onboard computer, while baseline1 occasionally triggered timeout warnings.

\begin{table}[b!]
\centering
\caption{The onboard processing time comparison.}
\begin{tabular}{lc}
\toprule
& \bf{Processing time per frame (ms)}\\
\midrule
Baseline1 & $34.10 \pm 4.13$ \\
Baseline2 & $2.43 \pm 0.17 $ \\
VAE-based (proposed) & $2.85 \pm 0.19$ \\
\bottomrule
\end{tabular}
\label{tab:process_time}
\end{table}

\begin{figure}[t]
    \centering
    \includegraphics[width=0.75\linewidth]{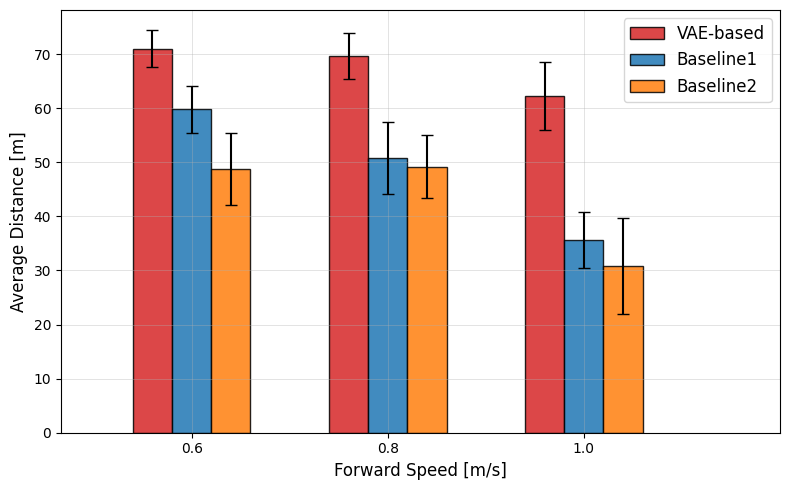}
    \caption{Average distance traveled before a failure under different forward speeds in the mix-species orchard.}    
    \label{fig:speed_change}
\end{figure}

We also evaluated the performance of the controllers under varying forward speeds. Since all training was conducted at a constant forward speed of 0.6 m/s, we increased the forward speed to 0.8 m/s and 1.0 m/s in the generalization environments to assess how the controllers would perform. Baseline2 (pre-trained) was excluded from this evaluation because its human intervention rate was too high, even at the nominal speed. While testing at even higher speeds is possible, it would have left the pilot with insufficient time to execute recovery maneuvers, so we limited the speed tests to 1.0 m/s. We used models trained over three iterations and flew the drone in two generalization environments, conducting two flights in each environment, resulting in four trails at each speed. The results, presented in Figure~\ref{fig:speed_change}, show the average distance flown by the drone without failure at different speeds. As expected, more failures occurred at higher speeds, reducing the distance traveled and increasing the need for human intervention. However, we found that our VAE-based controller generalized well to speed changes and achieved better results compared to the baselines. These results demonstrate that our VAE-based is robust to speed variations and delivers strong performance even when it encounters speeds outside its training range.

\section{Discussion}
\label{sec:discussion}

In our system, the EKF, altitude controller, and velocity controller each play a critical and practically indispensable role in ensuring robust and safe UAV flight. The EKF provides accurate state estimation for stable navigation and is a standard component in autonomous robotic systems. The altitude and velocity controllers maintain consistent flight conditions, which helps reduce variability introduced by human pilots during demonstration collection. This consistency is particularly important in imitation learning, as it allows the policy to focus on learning yaw control without being affected by fluctuations in speed or altitude. Additionally, maintaining constant altitude and velocity aligns with typical requirements in precision agriculture, where UAVs are expected to monitor crops at specific heights and speeds. Therefore, these components are integral to our system design and are essential for supporting the reliable training and evaluation of the proposed VAE-based control policy.

Our approach does not require explicit planning-control task decomposition and, therefore, saves a lot of resources and computational time. It directly learns a reactive visuomotor policy from human expert demonstrations, reliably controlling a drone navigating in orchards using only a single forward-looking camera. Based on the experimental results, the VAE-based controller effectively navigated the drone and avoided collisions after just three iterations of training. In this section, we discuss why our VAE-based controller may outperform the existing baselines.

\begin{figure}[b!]
    \centering
    \subfigure[]{
    \includegraphics[width=0.48\linewidth]{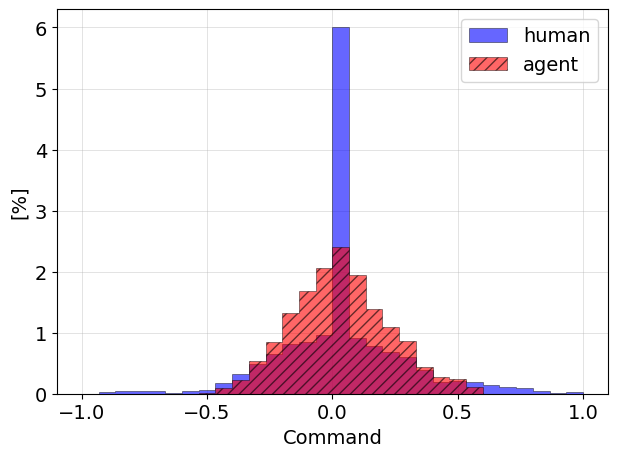}
    \label{fig:regression_histogram}
    }
    \subfigure[]{
    \includegraphics[width=0.48\linewidth]{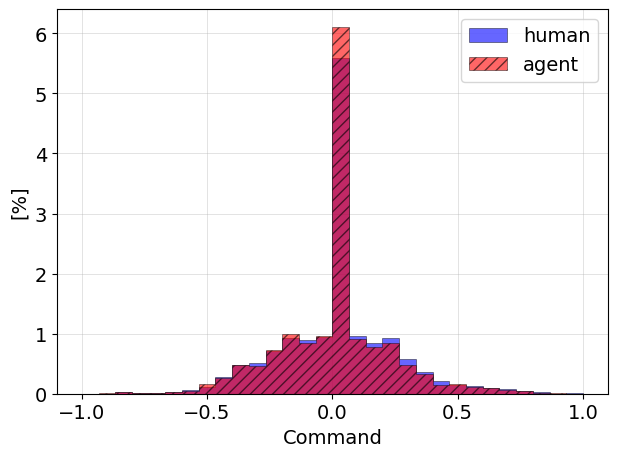}
    \label{fig:end2end_histogram}
    }
    \subfigure[]{
    \includegraphics[width=0.48\linewidth]{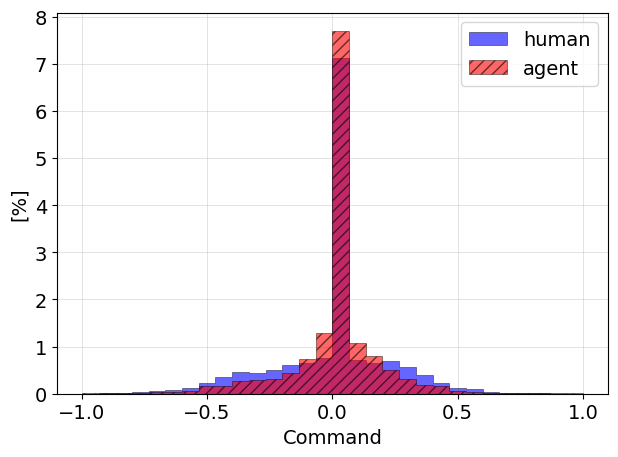}
    \label{fig:vae_histogram}
    }
    \caption{Histogram of human demonstration compared to the model prediction. The horizontal axis is the normalized control command and the vertical axis is the probability density. The model prediction comes from (a) baseline1 controller, (b) baseline2 controller, and (3) VAE-based controller (ours).}
    \label{fig:command_histogram}
\end{figure}

The histograms of human demonstrations compared to agent predictions are plotted in Figure~\ref{fig:command_histogram}. The horizontal axis represents the normalized control command, while the vertical axis shows the probability density of human pilot commands against the model-predicted commands from the training dataset. Regarding the baseline2 controller, we observed that its distribution closely matches the human distribution, indicating overfitting to the training data. This explains why the baseline2 controller struggles to generalize to novel environments when the distributions differ. In contrast, the baseline1 controller, which uses a linear regression method to calculate weights, produces a more conservative distribution, with predictions concentrated in the middle range of the control space. This hampers the baseline1 controller’s ability to generate large yaw commands when needed. The VAE-based controller, however, balances the distribution variance and fitting performance, allowing it to outperform the other two methods.

\begin{figure}[t!]
    \centering
    \subfigure[]{
    \includegraphics[width=0.45\linewidth]{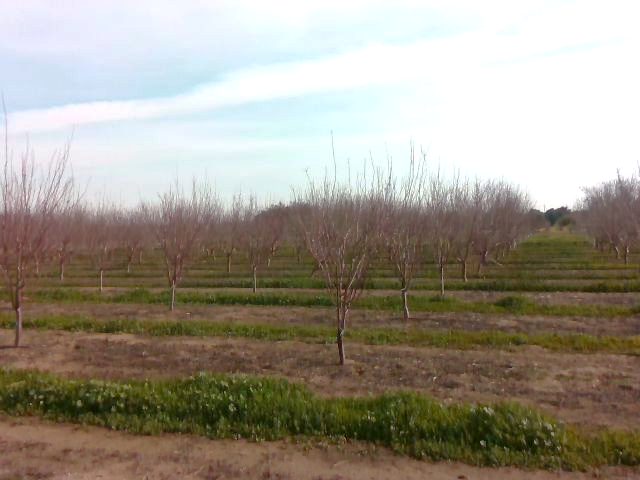}
    \label{fig:missing_tree}
    }
    \subfigure[]{
    \includegraphics[width=0.45\linewidth]{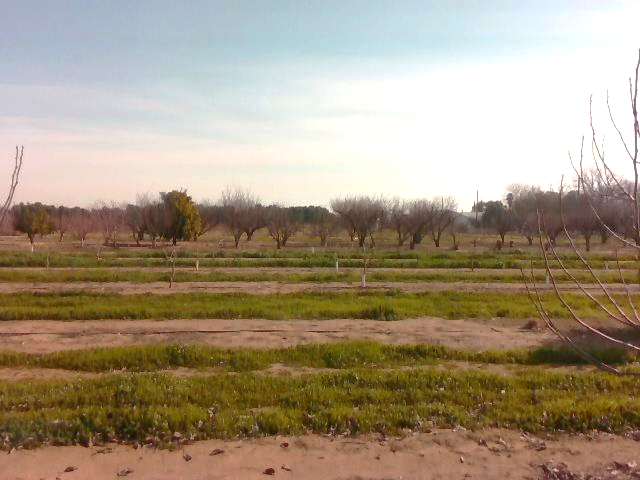}
    \label{fig:lack_of_feature}
    }
    \caption{Locations where the policy is more likely to fail (a) missing trees on the side; and (b) at the end of the row.}
    \label{fig:failure_locations}
\end{figure}

The VAE-based controller demonstrated strong autonomous navigation in the field, correctly avoiding obstacles. However, it did encounter challenges in specific situations. We observed that the controller was more likely to fail when some trees were missing on one side of the row (Figure~\ref{fig:missing_tree}), causing the orientation of the path to become ambiguous for the drone. In these cases, human intervention was required to re-orient the drone. A second failure occurred near the end of the row (Figure~\ref{fig:lack_of_feature}), where visual features for navigation became sparse, making the policy output unreliable. We had to switch to manual control when the drone crossed the last pair of trees. These failure cases will be addressed in future work. Possible solutions include adding memory to the agent to stabilize heading commands in the missing tree scenario and switching to a GPS-based planner at the end of the row to guide the agent and smoothly transition between rows.

Ideally, performance should improve continuously as more data are collected over multiple iterations. However, we observed challenges when the human pilot struggled to judge whether the agent’s decisions would lead to a collision, especially as the agent policy became more competent. In these cases, the pilot tended to wait until the agent was close to crashing before intervening, leading to inconsistent data. Retraining with such data could degrade the agent’s performance, potentially making the learning process less efficient. From our field experiments, we empirically determined that human pilots were able to provide high-quality demonstrations for up to three iterations. Consequently, we terminated our training after three iterations. We acknowledge the challenges that arise when a human teacher is part of the learning loop, as human inconsistency and errors can affect training quality. To improve consistency, a computer-aided module could be added to assist with intervention decisions.

Currently, our trained policy cannot be deployed “zero-shot” on platforms with different sizes and dynamics. This is because the human demonstrations were collected on a specific drone, and the same control inputs may not apply to drones with different masses or inertia. One possible solution is to integrate a robust low-level controller with an adaptive module, similar to the approach in \citet{zhang2023learning}, which would facilitate easier transfer of the controller to other drones during deployment.

\section{Conclusion and Future Work}
\label{sec:conclusion}

In this work, we presented a vision-based navigation policy that enables a UAV to autonomously fly within orchard rows. The policy utilizes a variational autoencoder neural network to extract latent information from image data captured by a front-mounted camera, generating reactive yaw rate commands for precise navigation. Our intervention-based learning approach leverages human expertise to guide the learning process, with pilots providing real-time corrective interventions during rollout. This method efficiently demonstrates safe behaviors and facilitates incremental learning. We validated the performance of our navigation policy in a real orchard, where the results showed that our VAE-based controller allowed the UAV to cover longer distances with less human assistance compared to existing baselines. Additionally, the controller exhibited strong generalization capabilities, adapting well to novel environments and speed changes. These results highlight the potential of our approach to enhance precision agriculture practices by enabling more efficient and autonomous UAV operations in complex orchard settings.

Future work will focus on addressing the limitations of our current system, particularly the need for human operators to monitor and intervene when the agent encounters unsafe behaviors. To eliminate this dependency, we aim to design a recovery planning module that will enable the agent to autonomously recover from unsafe states and continue its mission. Additionally, we will incorporate low-level control and account for UAV dynamics, making the policy transferable to other platforms of different sizes and dynamics. We also plan to investigate common failure cases to further enhance the controller’s robustness and evaluate its performance in dynamic environments with moving objects. Although this study mainly focuses on single-UAV navigation, the proposed framework can be scaled to support multiple UAVs operating simultaneously. Typically, UAVs could be distributed across orchard regions to maximize coverage and reduce collision risks. When multiple UAVs operate closely in the same region, a coordination planner could manage their interactions by enabling UAVs to share position and orientation data with neighboring UAVs, thus proactively adjusting trajectories to avoid collisions. To coordinate effectively, a game theory–based policy could be incorporated by modeling UAV interactions as potential games. Additionally, multi-agent reinforcement learning could be used to learn and optimize swarm behaviors through cooperative training in simulated environments.

\section*{CRediT authorship contribution statement}
\textbf{Peng Wei}: Writing – review \& editing, Writing – original draft, Visualization, Validation, Software, Methodology, Investigation, Formal analysis, Data curation, Conceptualization. \textbf{Prabhash Ragbir}: Writing - review \& editing, Validation, Investigation, Data curation. \textbf{Stavros G. Vougioukas}: Writing – review \& editing, Supervision, Resources, Funding acquisition, Conceptualization. \textbf{Zhaodan Kong}: Writing - review \& editing, Project administration, Supervision, Resources, Funding acquisition, Methodology, Conceptualization.

\section*{Declaration of competing interest}
The authors declare that they have no known competing financial interests or personal relationships that could have appeared to influence the work reported in this article.

\section*{\normalsize Data availability} \vspace{-3mm}
The dataset generated during the current study is available at: \url{https://drive.google.com/drive/folders/1tZiZu2b680ZqZX4IVkYo1QmVzZMHSpVP?usp=drive_link}.

\section*{Acknowledgements}
This work was supported by the USDA/NIFA AI Institute for Next Generation Food Systems (AIFS) under award 2020-67021-32855 and the Office of Naval Research (ONR) under NEPTUNE 2.0 award N00014-20-1-2268.


\bibliography{reference}

\end{document}